\pdfoutput=1
\documentclass[11pt]{article}
\usepackage[preprint]{acl}
\usepackage[T1]{fontenc}
\usepackage[utf8]{inputenc}
\usepackage{times}
\usepackage{latexsym}
\usepackage{tabularx}
\usepackage{graphicx}
\usepackage{hyperref}
\usepackage{fix-cm}
\usepackage{threeparttable}
\usepackage{fontawesome5}
\usepackage{url}
\usepackage{color}
\usepackage{longtable}
\usepackage{amsmath}
\usepackage{booktabs}
\usepackage{makecell}
\usepackage{float}
\usepackage{subfigure}
\usepackage{multirow}
\usepackage{wrapfig}
\usepackage{inconsolata}
\usepackage{bbding}
\usepackage{arydshln}
\usepackage{lipsum} 
\usepackage{microtype}
\usepackage{mdframed}

\title{Self-Steering Optimization: Autonomous Preference Optimization for Large Language Models}

\vspace{5px}
\author{
\hspace{-2px}Hao Xiang${}^{1,3}$,
Bowen Yu${}^{2}$\thanks{~ Corresponding authors.},
Hongyu Lin${}^{1}$\footnotemark[1],
Keming Lu${}^{2}$, 
Yaojie Lu${}^{1}$,
\\
\textbf{Xianpei Han}${}^{1}$,
\textbf{Ben He}${}^{1,3}$,
\textbf{Le Sun}${}^{1}$,
\textbf{Jingren Zhou}${}^{2}$, 
\textbf{Junyang Lin}${}^{2}$ 
\vspace{5px}
\\
$^{\rm 1}$Chinese Information Processing Laboratory, Institute of Software, \\Chinese Academy of Sciences \\
$^{\rm 2}$Alibaba Group  \hspace{10px} $^{\rm 3}$University of Chinese Academy of Sciences \vspace{5px}  \\
\hspace{1px}\{xianghao2022, hongyu, luyaojie, xianpei, sunle\}@iscas.ac.cn  \hspace{5px} {benhe@ucas.edu.cn}  \\ \hspace{1px}\{yubowen.ybw, lukeming.lkm, jingren.zhou, junyang.ljy\}@alibaba-inc.com
}

\begin{document}

\maketitle
\begin{abstract}
The key to effective alignment lies in high-quality preference data. Recent research has focused on automated alignment, which involves developing alignment systems with minimal human intervention.
However, prior research has predominantly focused on developing data generation methods, while insufficient attention has been paid to quality control mechanisms, which often produce inaccurate and unhelpful data, leading to unpredictable benefits during iterative optimization.
In this paper, we present Self-Steering Optimization ($SSO$), an algorithm that autonomously generates high-quality preference data, eliminating manual annotation requirements.
$SSO$ employs a specialized optimization objective to build a data generator from the policy model itself, which is used to produce accurate and on-policy data.
We demonstrate $SSO$'s effectiveness through comprehensive experiments on two series of models: Llama 3 and Qwen 2. Our evaluation across diverse benchmarks shows that $SSO$ consistently outperforms baselines in human preference alignment and reward optimization. Further analysis validates $SSO$ as a scalable framework for preference optimization, benefiting the advancement in automated alignment techniques.

\end{abstract}

\section{Introduction}
\begin{figure*}[ht!]
  \centering
  \subfigure[Ideal preference data.]{
    \includegraphics[width=0.3\textwidth]{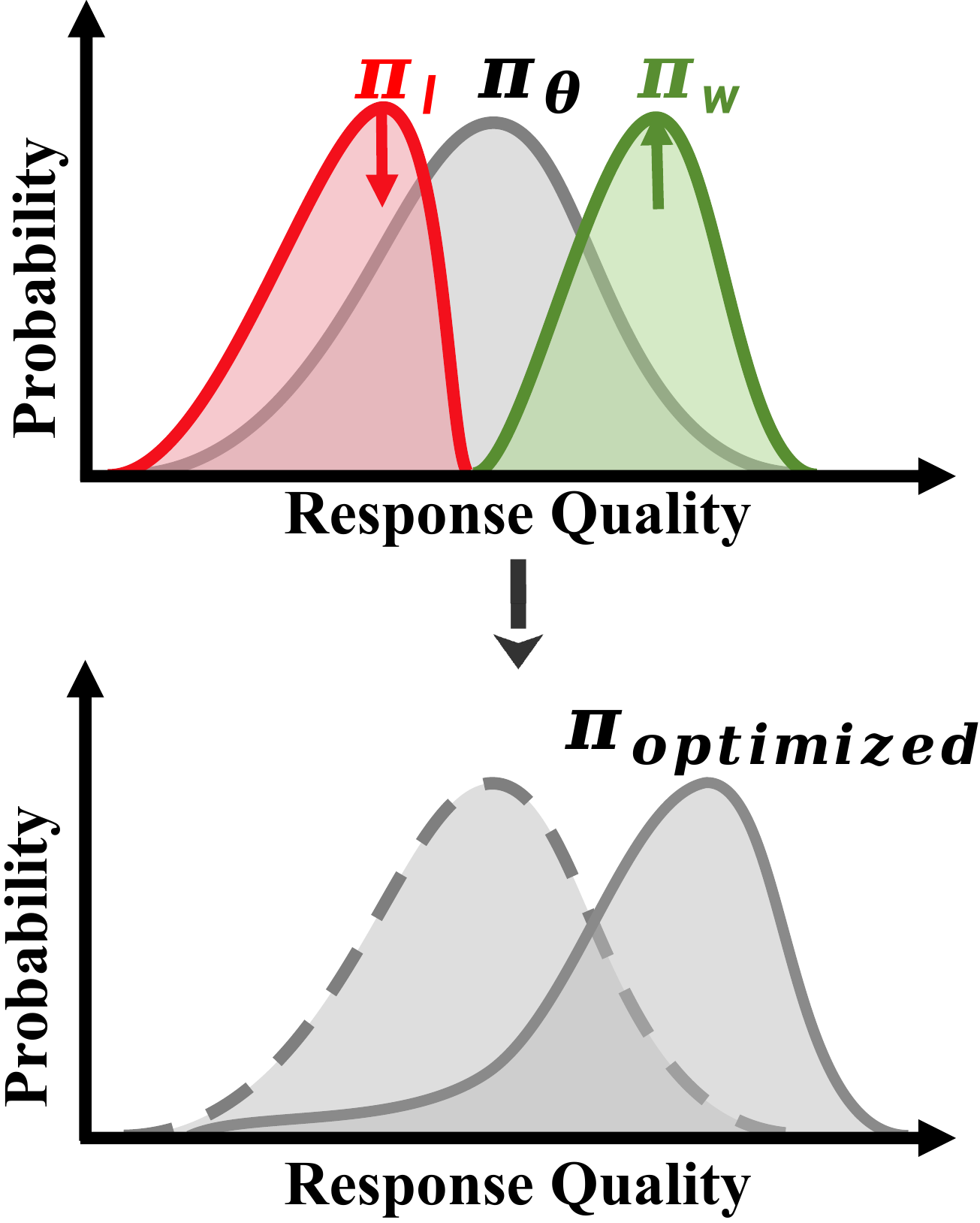}
    \label{dist_G}
  }
  \hspace{-10pt}
  \subfigure[Accurate but off-policy data.]
  {
    \includegraphics[width=0.3\textwidth]{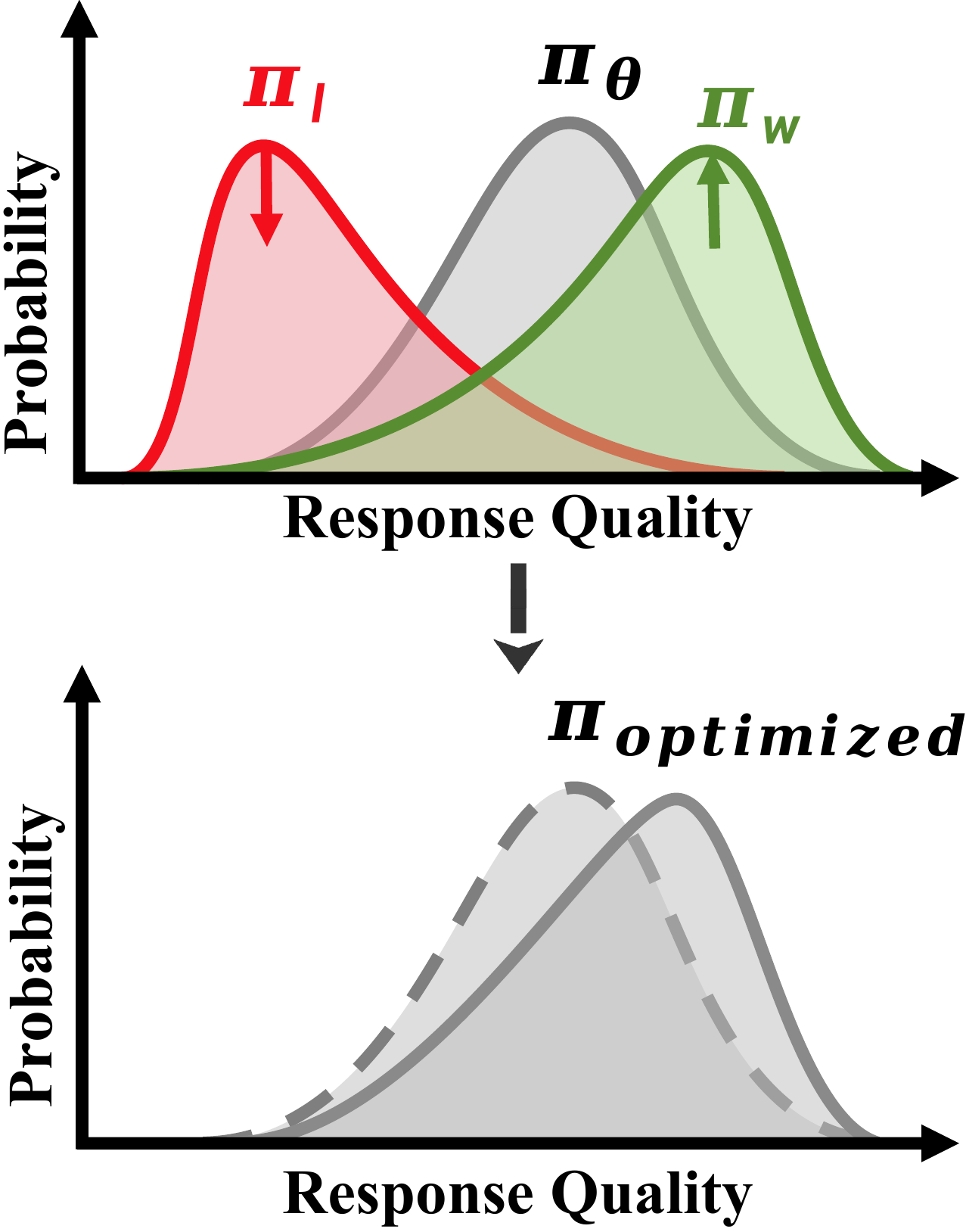}
    \label{dist_A}
  } 
  \hspace{-10pt}
  \subfigure[On-policy but inaccurate data.]
  {
    \includegraphics[width=0.3\textwidth]{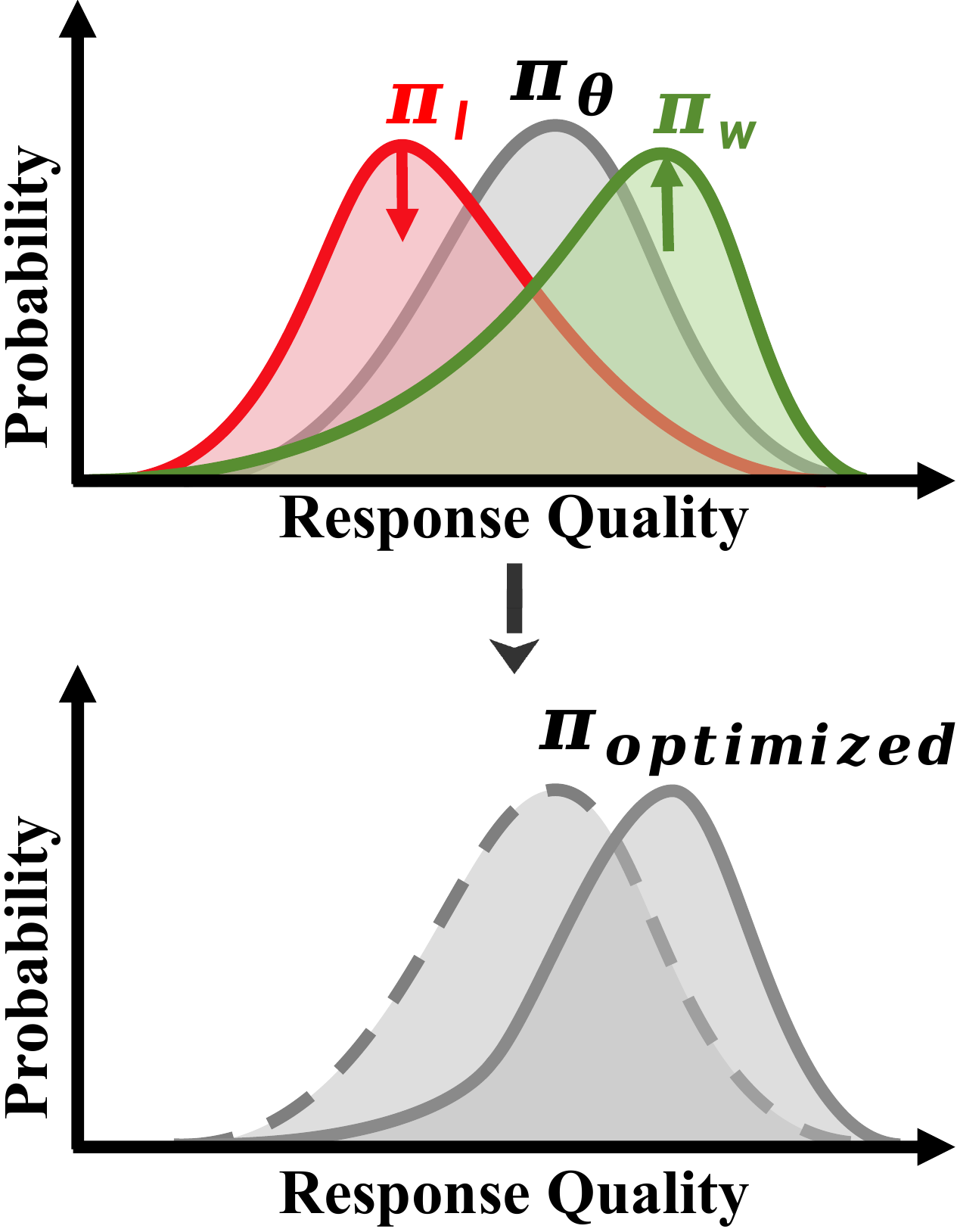}
    \label{dist_O}
  } 
  \caption{Distribution changes of policy model $\pi_{\theta}$ after optimization with three types of preference data.
During alignment, chosen responses $\pi_{w}$ receive positive gradients to increase probability, while rejected responses $\pi_{l}$ receive negative gradients to decrease probability.
(a) Ideal data, with high accuracy (low overlap between $\pi_{w}$ and $\pi_{l}$) and on-policy nature ($\pi_{l}$ lies in high-probability region of $\pi_{\theta}$), leading to right optimization direction and effective negative gradients optimization.
(b) Sub-optimal data, with high accuracy (low overlap between $\pi_{w}$ and $\pi_{l}$) and off-policy nature ($\pi_{l}$ lies in low-probability region of $\pi_{\theta}$), weakening negative gradients optimization.
(c) Sub-optimal data, with low accuracy (high overlap between $\pi_{w}$ and $\pi_{l}$) and on-policy nature ($\pi_{l}$ lies in high-probability region of $\pi_{\theta}$), interfering optimization direction.
}
\label{moti}
\end{figure*}

The field of Natural Language Processing has undergone revolutionary advancements driven by large language models (LLMs). After meticulous alignment processes, LLMs have demonstrated remarkable capabilities in following instructions and understanding human preferences. This leads to the development of widely acclaimed products like ChatGPT~\citep{openai_chatgpt}, which have captured significant public attention.
However, aligning LLMs with human preferences is not trivial. 
Despite the existence of Proximal Policy Optimization (PPO)~\citep{ouyangTrainingLanguageModels2022b}, an ideal alignment training process requires a robust reward model and a stable reinforcement learning process, encouraging researchers to develop offline preference optimization algorithms such as Direct Preference Optimization (DPO)~\citep{rafailovDirectPreferenceOptimization2023a}.
However, algorithms like DPO rely on a substantial amount of high-quality, annotated preference data, which is both resource-intensive and requires meticulous attention. In addition, the limited capabilities of human annotators cause inherent limitations in annotated data, making it challenging to achieve \textit{superalignment}~\citep{burns2023weaktostrong}.

Consequently, recent researchers have shifted their focus towards automated alignment, with the intention of developing scalable, high-quality alignment systems with minimal human intervention. The cornerstone of this paradigm is the pursuit of scalable alignment signals that are capable of effectively replacing human-annotated preference data. Current popular strategies include self-judgement~\citep{yuan2024self, wu2024metarewarding}, principle-based automated alignment~\citep{yang2024rlcd, bai2022constitutional}, Constitutional AI~\citep{bai2022constitutional}, and other methods~\citep{franken2024selfsupervised, kumar2024traininglanguagemodelsselfcorrect}.

However, these methods do not pay enough attention to quality control mechanisms. The ideal preference data, as defined in RLHF \citep{ouyangTrainingLanguageModels2022b}, is constructed through human annotation, where the responses are sampled from the policy model and ranked according to their qualities. As shown in Figure \ref{dist_G}, this type of preference data demonstrates high accuracy\footnote{\textbf{Accuracy of preference data}: the rate of the response pairs where the chosen response has higher quality than the rejected response.} and on-policy nature\footnote{\textbf{On-policy data}: the responses lie in the high-probability region of the policy model. The higher generation probability indicates better on-policy behavior.}, leading to ideal alignment optimization~\cite{tajwar2024preference, kim2024evaluatinglanguagemodelssynthetic}. However, automated methods often fail to simultaneously guarantee accurate preferences and on-policy responses, instead producing suboptimal data, as shown in Figures \ref{dist_A} and \ref{dist_O}, which impedes model alignment. For example, self-judgment is hampered by the inherent limitations of the model; this judging ability is restricted and difficult to improve, often resulting in hacked rewards and inaccurate preference data in Figure \ref{dist_O}~\citep{wu2024metarewarding}. In other methods, incorporating additional input or processes may lead to off-policy responses and sub-optimal preference data, as shown in Figure \ref{dist_A}.

We then recognized the need for a novel approach to generate high-quality preference data to address these limitations and advance automated alignment. One problem is to control the distribution of the chosen and rejected responses when constructing preference data. For most automated alignment methods, this seems an impossible task because chosen and rejected responses are typically obtained through complex pipelines. However, we found that principle-based methods \citep{yang2024rlcd, bai2022constitutional} can achieve this goal because they construct preference data by directly sampling from the policy model based on good and bad principles. It can be approximated that the distributions of the chosen and rejected responses $\pi_{w}$ and $\pi_{l}$ are the distribution of the policy model $\pi_{\theta}$ with good and bad principles.

In this work, we introduce Self-Steering Optimization ($SSO$), a pioneering method that automatically generates accurate and near-on-policy preference data for the policy model. $SSO$ optimizes a data generator with a special loss to control the distributions of chosen and rejected responses, and then uses the data produced by this generator to further optimize the policy model. Specifically, $SSO$ first prompts the policy model with original queries $x$ and a set of contrastive principles $p^+$ and $p^-$ for the responses as training data and then optimizes the policy model based on two key objectives:
a) Making rejected responses approximately on-policy to ensure the effectiveness of negative gradients in subsequent optimization. We only take care of the rejected responses, as the chosen responses usually have high generation probabilities.
b) Maintaining a consistent gap between the chosen and rejected responses to ensure the accuracy of the preference data.

We demonstrate the effectiveness of Self-Steering Optimization on Qwen2~\citep{yang2024qwen2technicalreport} and Llama3~\citep{dubey2024llama3herdmodels} backbones.
Our experiments reveal $SSO$'s ability to generate accurate and on-policy preference data. As a result, improvements are observed on a wide range of benchmarks such as MATH~\citep{hendrycksmath2021}, IFEval~\citep{zhou2023instructionfollowingevaluationlargelanguage}, MT-Bench~\citep{zheng2024judging}, and AlpacaEval 2.0~\citep{dubois2024lengthcontrolledalpacaevalsimpleway}. Furthermore, we conducted experiments through reward optimization, which also achieved satisfying results.
Without human annotation or external models, $SSO$ even outperforms baselines with annotated data~\citep{cui2024ultrafeedback}, underscoring its potential as a scalable and efficient alignment approach.

\section{Related Works}
\paragraph{Preference Alignment}
Researchers have proposed various algorithms to align large language models (LLMs) with human preferences. 
\citet{ziegler2020finetuninglanguagemodelshuman, ouyangTrainingLanguageModels2022b, bai2022traininghelpfulharmlessassistant} train a reward model based on annotated human preference data and employ reinforcement learning algorithms such as PPO \citep{schulman2017proximalpolicyoptimizationalgorithms} to align LLMs. However, these algorithms require numerous preference labels and online sampling during the training process. To further reduce costs, direct preference optimization (DPO), sequence likelihood calibration (SLiC)~\citep{zhao2023calibrating}, identity preference optimization (IPO)~\citep{azar2023generaltheoreticalparadigmunderstand}, and Kahneman-Tversky optimization (KTO)~\citep{ethayarajh2024ktomodelalignmentprospect} simplify the RLHF objective by directly increasing the margin between chosen and rejected responses. 

\paragraph{Automated Alignment}
\label{autoalign}
Previous alignment studies rely on manually annotated preference data and algorithms such as RLHF and DPO to conduct model alignment. Recently, numerous studies have found that LLM-generated data can reach the quality of ordinary manual annotations \citep{zheng2024judging}. These findings increased the attention of automated alignment \citep{yuan2024self,chen2024self}. Automated alignment aims to minimize human intervention by addressing the prohibitively expensive cost of human annotation. Current methods can be divided into four types based on the source of alignment signals~\citep{cao2024scalableautomatedalignmentllms}: 1) Inductive Bias, from introducing appropriate assumptions and constraints~\citep{huang2023large, bai2022constitutional, yang2024rlcd, yuan2024self, chen2024self}. 2) Behavioral Imitation, another aligned model~\citep{peng2023instruction, tunstall2023zephyr, burns2023weaktostrong}. 3) Model Feedback, feedbacks from other models~\citep{lee2023rlaif, hosseini2024vstar}. 4) Environmental Feedback, environmental interaction~\citep{liu2023training, qiao2024making}.
\section{Preliminaries}
\subsection{Symbol Definition of Automated Alignment}
Specifically, given a query set $X=\{x_i\}^{N}_{i=1}$, where $N$ is the number of queries, automated methods focus on how to use the policy model $\pi_{\theta}$ to generate the chosen response $y^+$ and the rejected response $y^-$ for the preference data $D=\{x_i, y^+_i, y^-_i\}^{N}_{i=1}$, which will be used to optimize $\pi_{\theta}$ with alignment algorithms.

\subsection{Principle-Based Automated Alignment}
Principle-based automated alignment (PBAA) is one of the most common automated alignment methods~\citep{yang2024rlcd,franken2024selfsupervised}, which assumes that responses with different qualities can be directly sampled from LLMs with different queries. This approach constructs pairs of contrastive queries $x^+$ and $x^-$ to sample chosen and rejected responses from the policy model as training data. Since contrastive queries exhibit contrasting attributes (such as harmful vs. harmless), the generated preference data has high accuracy. Representative works of PBAA include RLCD~\citep{yang2024rlcd}, AutoPM~\citep{huang-etal-2023-learning-preference} and SAIM~\citep{franken2024selfsupervised}. The first two use specific word pairs, such as "inoffensive response" and "offensive response", to generate response pairs for model alignment, while SAIM employs generated principles.

However, these methods do not guarantee accurate and on-policy data. Incorporating additional principles could lead to off-policy responses, as shown in Figure \ref{dist_A}~\cite{tajwar2024preference,kim2024evaluatinglanguagemodelssynthetic}.

\begin{figure*}[ht]
\begin{center}
\includegraphics[width=\textwidth]{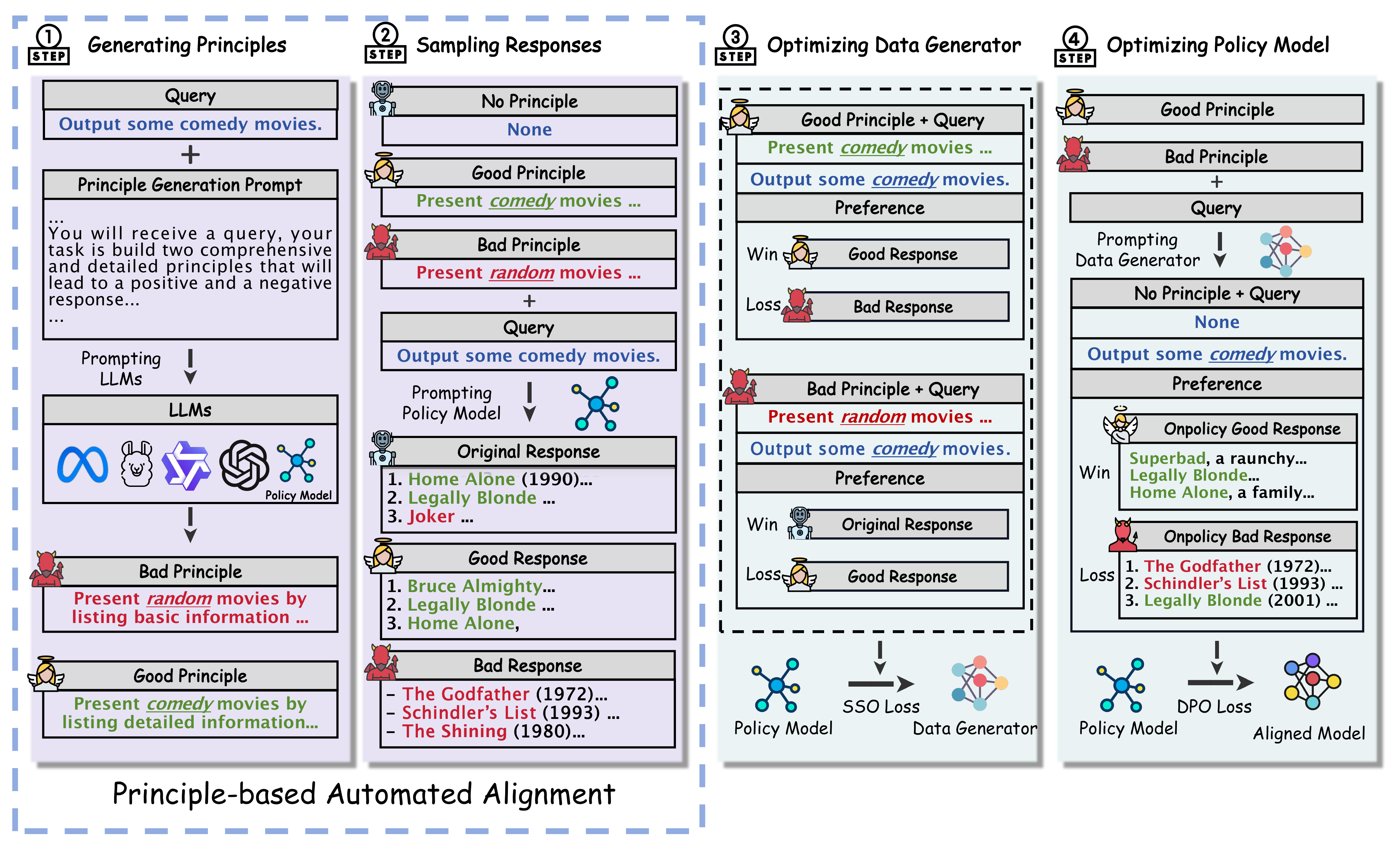}
\end{center}
\caption{
    The pipeline of Self-Steering Optimization ($SSO$). The overall process consists of four steps: 1) generate Principles, 2) sample Responses, 3) optimize the Data Generator, and 4) optimize the policy model. The data generator is optimized with the loss \ref{loss} from the policy model $\pi_{\theta}$ and used to generate accurate and on-policy preference data for the policy model.
}
\label{pdpo}
\end{figure*}

\subsection{Automated Alignment from Distribution Sight}
\label{motivation}
In the alignment process, negative gradients and positive gradients are used to decrease and increase the probability of rejected responses and chosen responses, respectively. Figure \ref{dist_G} illustrates the ideal distribution of the preference data.

For models that are not aligned at all, automated alignment approaches may construct preference data where the distribution of the chosen responses $\pi_w$ is far from the distribution of rejected responses $\pi_l$, and $\pi_w$ are in the low-probability region of $\pi_{\theta}$, as mentioned by \citet{tajwar2024preference}, the on-policy nature has minimal impact on model optimization in this scenario. This explains the improvements of various automated methods.

However, as LLMs advance, even those without explicit alignment can exhibit aligned behaviors. In this situation, $\pi_w$ lies in the high-probability region of $\pi_{\theta}$, and the on-policy performance of $\pi_l$ becomes crucial. If $\pi_l$ lies in the low-probability region of $\pi_{\theta}$, applying negative gradients to such responses would be meaningless and result in sub-optimal alignment. Therefore, we propose Self-Steering Optimization.
\section{Self-Steering Optimization}
\subsection{Pipeline of SSO}
\label{implementation}
As shown in Figure \ref{pdpo}, the pipeline of $SSO$ consists of four steps: 1) generating principles, 2) sampling responses, 3) optimizing the data generator, and 4) optimizing the policy model.
\paragraph{Generating Principles} Given a query $x$, we employ a principle generator (we used Qwen2.5-72B-Instruct\citep{qwen2025qwen25technicalreport} in our main experiments) to construct a pair of contrastive principles ($p^+$, $p^-$). The example of principles can be found in the Appendix \ref{principle}.
\paragraph{Sampling Responses} Principles ($p^+$, $p^-$) are then concatenated with the original query $x$ to build contrastive queries ($x^+$, $x^-$).\footnote{We use principles as system messages.} Then, ($x^+$, $x^-$) are used to prompt the policy model $\pi_{\theta}$ for the good and bad responses ($y^+$, $y^-$). Furthermore, to optimize the data generator, we generate an original response $y^o$ without any principle.
\paragraph{Optimizing Data Generator} We optimize a data generator with the loss function in Formula \ref{loss}, which is designed to generate accurate and near-on-policy preference data. The data generator is optimized from the policy model.
\paragraph{Optimizing Policy Model} We optimize the policy model $\pi_{\theta}$ with the preference data generated by the data generator. The policy model is optimized with an alignment algorithm, such as DPO, to align the model. 

Detailed templates are provided in the Appendix.

\subsection{Objective}
\label{objective}
Self-Steering Optimization aims to generate accurate and near-on-policy preference data.
As described in Section \ref{implementation}, given the following components:

\paragraph{Two principles}: Good principle $p^+$ and bad principle $p^-$, which are generated by LLMs for the query $x$ specially.

\paragraph{Three queries}: Original query $x$, good query $x^+$, and bad query $x^-$, where $x^+$ and $x^-$ are built from $x$ with $p^+$ and $p^-$, respectively.

\paragraph{Three responses}: The good response $y^+$, the bad response $y^-$, and the original response $y^o$, where $y^i$ is sampled from $\pi_{\theta}$ with query $x^i$, $i\in\{+, -, o\}$.

We propose a novel loss function $\mathcal{L}_{SSO}$ to optimize the data generator:
\begin{equation}
\label{loss}
\mathcal{L}_{SSO} = \mathcal{L}^{+}+\mathcal{L}^{-}
\end{equation}
where $\mathcal{L}^{+}$ and $\mathcal{L}^{-}$ are used to optimize the good and bad responses, respectively. 

Loss \ref{loss} is the core loss function of our method, used to optimize the Policy model to obtain a Data Generator (as shown in Step 3 of Figure 1). Although the Data Generator is optimized from the Policy Model, it has no direct relationship with the final aligned model and is only used to generate data. $\mathcal{L}_{SSO}$ is only used to optimize the Data Generator, not for the final alignment. When aligning the policy model with the data generated by the Data Generator, we use $\mathcal{L}_{Base}$, which is the $xPO$ Loss, such as DPO, IPO, etc. This might be the source of your confusion.

As mentioned in Section \ref{motivation}, $\mathcal{L}_{SSO}$ should minimize the overlap between $\pi_{w}$ and $\pi_{l}$, while ensuring that $\pi_{l}$ lies in the high-probability region of $\pi_{\theta}$ (for effective negative gradients optimization). Naturally, we design $\mathcal{L}^{+}$ as:
\begin{equation}
    \mathcal{L}^{+} = L_{Base}(\mathbf{x^+},\mathbf{y^+},\mathbf{y^-}) + \gamma L_{sft}(\mathbf{x^+},\mathbf{y^+})
\end{equation}
where $L_{Base}$ can be any $xPO$ alignment algorithm(like DPO, IPO, etc.) loss function, and $L_{sft}$ is the SFT loss function. $\gamma$ is a hyperparameter that balances SFT loss and alignment loss, helping to maintain training stability.

Similarly, a natural approach is to construct the loss function $\mathcal{L}^{-}$ as:
\begin{equation}
    \label{rloss1}
    \mathcal{L}^{-} = L_{Base}(\mathbf{x^-},\mathbf{y^-},\mathbf{y^+}) + \gamma L_{sft}(\mathbf{x^-},\mathbf{y^-})
\end{equation}
However, this approach introduces a problem: with a bad principle $p^-$, LLMs may output unpredictable results. In other words, this loss could lead to a $\pi_{l}$ that lies in the low-probability region of $\pi_{\theta}$, which cannot help generate on-policy data and hamper alignment.

Therefore, for the optimization of $\pi_{l}$, we change the loss to $L_{Base}(\mathbf{x^-},\mathbf{y^o},\mathbf{y^+})$. This goal is crucial, as we want to avoid shifting $p^-$ to the low-probability region of $\pi_{\theta}$. And the final form of $\mathcal{L}^{-}$ is:
\begin{equation}
    \label{rloss2}
    \mathcal{L}^{-} = L_{Base}(\mathbf{x^-},\mathbf{y^o},\mathbf{y^+}) + \gamma L_{sft}(\mathbf{x^-},\mathbf{y^o})
\end{equation}
\section{Experiments}
\begin{table*}[!t]
	\setlength\tabcolsep{1.5pt}
	\centering
		\fontsize{11}{10}
		\selectfont
          \resizebox{\textwidth}{!}{
		\begin{tabular}{lccccccccc}
			\toprule
			\multirow{2}{*}{\makecell[c]{Model}} & \multirow{2}{*}{\makecell[c]{Synthetic\\Data}} & \multicolumn{2}{c}{\makecell[c]{AlpacaEval}}  & MTbench & IFEval & GSM8K & MATH  & MMLU \\
            \cmidrule(lr){3-10}
 &            & lc win rate & win rate & score & avg score & acc  & acc   & acc  \\
			\midrule
    \multicolumn{2}{l}{\textsc{Llama3-SFT}~\citep{InfinityInstruct2024}} &20.6  & 15.0 & 7.38  & 24.9  & 75.6  & 29.5  & 65.9  \\
    w/ \textsc{UltraFeedback}~\cite{cui2024ultrafeedback} & \XSolidBrush & 22.0 & 17.5 & 7.71  & 43.6  & 78.3  & 30.5  & 66.4  \\
    \multicolumn{9}{l}{\textcolor{lightgray}{\textit{Principle-Based Automated Alignment}}} \\
    w/ \textsc{PBAA}$_{DPO}$ & \Checkmark & 29.5 & 24.0 & 7.92  & 47.8  & 77.9  & 30.4  & 66.3  \\
    w/ \textsc{SSO}$_{DPO}$ & \Checkmark & \textbf{\underline{35.0}} & \textbf{\underline{28.3}} & \textbf{\underline{7.96}} & \textbf{\underline{50.3}} & \textbf{\underline{80.5}} & \textbf{\underline{30.8}} & \textbf{\underline{66.7}} \\
   \noalign{\vskip 0.11cm}
    \hdashline
   \noalign{\vskip 0.11cm}
    \multicolumn{2}{l}{\textsc{Qwen2-SFT}~\citep{InfinityInstruct2024}} & 20.1 & 13.2 & 8.24  & 19.8  & 78.5  & 44.6  & \textbf{\underline{71.1}}  \\
    w/ \textsc{UltraFeedback} & \XSolidBrush & 20.2 & 15.0 & 8.35  & 40.4  & 84.3  & 46.7  & 71.1  \\
    \multicolumn{9}{l}{\textcolor{lightgray!99}{\textit{Principle-Based Automated Alignment}}} \\
    w/ \textsc{PBAA}$_{DPO}$ & \Checkmark & 37.7 & 42.5 & 8.59  & 43.6  & 83.8  & 50.8  & 70.9  \\
    w/ \textsc{SSO}$_{DPO}$ & \Checkmark  & \textbf{\underline{43.0}} & \textbf{\underline{45.4}} & \textbf{\underline{8.66}} & \textbf{\underline{45.7}} & \textbf{\underline{84.7}} & \textbf{\underline{52.3}} & 71.0 \\
   \noalign{\vskip 0.11cm}
    \hdashline
   \noalign{\vskip 0.11cm}
   \multicolumn{2}{l}{\textsc{Llama3-Instruct}~\citep{dubey2024llama3herdmodels}} &  20.1 & 13.2 & 8.06  & 53.0  & 80.4  & 28.5  & 68.4  \\
    \multicolumn{1}{l}{w/ \textsc{UltraFeedback}} & \XSolidBrush & 23.8 & 22.6 & 7.72  & \textbf{\underline{54.4}} & 79.1  & 29.6  & \textbf{\underline{68.4}} \\
    \multicolumn{9}{l}{\textcolor{lightgray!99}{\textit{Principle-Based Automated Alignment}}} \\
    w/ \textsc{PBAA}$_{DPO}$ & \Checkmark & 23.8 & 25.7 & 8.05  & 53.2  & 79.5  & 28.9  & 68.0  \\
    w/ \textsc{SSO}$_{DPO}$ & \Checkmark & \textbf{\underline{25.6}} & \textbf{\underline{28.9}} & \textbf{\underline{8.13}} & 53.4  & \textbf{\underline{80.7}} & \textbf{\underline{29.6}} & 68.4  \\
   \noalign{\vskip 0.11cm}
    \hdashline
   \noalign{\vskip 0.11cm}
    \multicolumn{2}{l}{\textsc{Qwen2-Instruct}~\citep{yang2024qwen2technicalreport}} & 19.5 & 17.2  & 8.33  & 51.4  & 81.0  & 40.0  & 71.0  \\
    \multicolumn{1}{l}{w/ \textsc{UltraFeedback}} & \XSolidBrush & 20.4 & 17.6  & 8.21  & \textbf{\underline{51.5}} & \textbf{\underline{82.3}} & 42.9  & 71.0  \\
    \multicolumn{9}{l}{\textcolor{lightgray!99}{\textit{Principle-Based Automated Alignment}}} \\
    w/ \textsc{PBAA}$_{DPO}$ & \Checkmark & 34.7 & 43.7 & 8.37  & 50.9  & 78.1  & 42.1  & 71.0  \\
    w/ \textsc{SSO}$_{DPO}$ & \Checkmark & \textbf{\underline{36.5}} & \textbf{\underline{49.4}} & \textbf{\underline{8.39}} & 51.4  & 78.5  & \textbf{\underline{44.2}} & \textbf{\underline{71.2}} \\
			\bottomrule  
		\end{tabular}
        }
\caption{Evaluation results on six distinct tasks. "lc win rate" indicates "Length Control Win Rate" from AlpacaEval 2.0 \citep{dubois2024lengthcontrolledalpacaevalsimpleway}.}
	\label{tab:main_tab}
	\vspace{-10pt}
\end{table*}
\subsection{Experimental Setup}
\label{expsetup}
\paragraph{Models and Datasets}
We conducted experiments primarily on Qwen2-7B~\citep{yang2024qwen2technicalreport} and Llama3-8B~\citep{dubey2024llama3herdmodels}. We used the SFT models from \citet{InfinityInstruct2024}, which were fine-tuned from the pretrain models with 3M data. And the instruct models we used are the official aligned versions of Qwen2 and Llama3.
For datasets, most of our experiments are based on UltraFeedback~\citep{cui2024ultrafeedback}. This dataset includes 60k annotations of preference data. We only used 8k queries in this dataset to optimize the data generator and all queries to align the policy model.

\paragraph{Training Setting}
We chose the DPO loss as the basic loss in the main experiments and also showed the results with the IPO loss~\citep{azar2023generaltheoreticalparadigmunderstand} in Section \ref{discussion}. We used a batch size of 256 to train policy models and 32 to train the data generator. We applied a simple hyperparameter search to determine the learning rate and the $\beta$ parameter in DPO. We trained models with the learning rate 5E-7 and $beta$ with 0.1. We set $\gamma$ in $SSO$ to $0.1$. We used the top-p = 0.8, temperature = 0.7, and max\_new\_tokens = 2048 for sampling responses. The training scripts were based on LlamaFactory\citep{zheng2024llamafactory} and RLHF Workflow\citep{dong2024rlhfworkflowrewardmodeling}.

\paragraph{Evaluation}
We evaluated the model performance on two widely used subjective evaluation benchmarks: MT-Bench~\citep{zheng2024judging} and AlpacaEval 2.0~\citep{dubois2024lengthcontrolledalpacaevalsimpleway}. MT-Bench comprises 80 questions with answers scored by GPT-4. AlpacaEval 2.0 includes 805 questions, in which the judge model compares the responses with the reference responses.
Additionally, we evaluated models on a series of objective benchmarks: MATH~\citep{hendrycksmath2021}, GSM8K~\citep{cobbe2021training}, MMLU~\citep{hendrycks2021measuringmassivemultitasklanguage}, and IFEval~\citep{zhou2023instructionfollowingevaluationlargelanguage}. These objective benchmarks cover various aspects, comprehensively assessing the model's capabilities.

\subsection{Main Results}
This part compares the performance of $SSO$ with $PBAA$ and UltraFeedback. 
Table \ref{tab:main_tab} demonstrates that $SSO$ achieved outstanding results on most benchmarks.

When optimizing the SFT model, $SSO$ showed an average improvement of nearly \textbf{14\%} on AlpacaEval 2.0 and \textbf{0.5} points on MTBench. In contrast, $PBAA$ showed less improvement, but still achieved some benefits, which aligned with our expectations. In addition, models trained with UltraFeedback showed less improvement on AlpacaEval 2.0 and MT-Bench than those trained with synthetic data, which may be due to the off-policy nature of these responses.
$SSO$ also showed benefits on objective benchmarks. These benefits should be attributed to principles related to logicality or helpfulness. Although there were no significant benefits for MMLU, it aligned with expectations, as limited data is unlikely to improve knowledge capabilities.
We also applied $SSO$ to aligned models such as Meta-Llama-3-8B-Instruct and Qwen2-7B-Instruct, with the results shown in Table \ref{tab:main_tab}. $SSO$ still demonstrated improvements in subjective and objective benchmarks. Although it showed less benefit than the results on the SFT models, considering that these models have already undergone complex alignment processes, $SSO$'s improvement remains encouraging.
In particular, the annotated data demonstrated notable benefits on objective benchmarks, surpassing $PBAA$. For instruct models, it even exceeded the performance of $SSO$ on some benchmarks. These results highlight the respective strengths and limitations of the synthetic data, aligning with the findings reported by \citet{shumailov2024ai}.

\subsection{Results in Reward Optimization}
\begin{table*}[!t]
	\centering
		\fontsize{10}{11.5}
		\selectfont
		\begin{tabular}{lccccccc}
			\toprule
			Training Data & Type  & Size  & Avg   & Chat  & Chat Hard & Safty & Reasoning \\
			\midrule
    \textsc{UltraFeedback} & Annotated & 60k   & 79.5  & 96.9  & 61.8  & 79.2  & 80.2  \\
    \textsc{PBAA}$_{DPO}$ & Synthetic & 60k   & 78.9  & 96.9  & 59.9  & 77.2  & 81.6  \\
    \textsc{SSO}$_{DPO}$ & Synthetic & 60k   & 80.0  & 95.3  & 59.0  & 77.4  & 88.3  \\ 
    \noalign{\vskip 0.09cm}
    \hdashline
   \noalign{\vskip 0.09cm}
    \multicolumn{8}{l}{\textcolor{lightgray!99}{\textit{Mixed with Skywork-Reward-Preference-80K-v0.2, a SOTA preference dataset, as training data.}}} \\
    \textsc{Skywork}~\cite{skyworkreward2024} & Annotated & 80k   & 84.5  & 91.6  & 78.6  & 88.2  & 79.7  \\   
    + \textsc{UltraFeedback} & Annotated & 140k  & 85.1  & 94.7  & 72.7  &  89.1  & 83.9  \\
    + \textsc{PBAA}$_{DPO}$ & Mixed   & 140k  & 83.5  & 93.3  & 75.1  & 86.5  & 79.0  \\
    + \textsc{SSO}$_{DPO}$ & Mixed   & 140k  &  \textbf{\underline{86.3}} & 93.9 & 75.4 & 86.6 & 89.3 \\
			\bottomrule  
		\end{tabular}
        \caption{Evaluation results on RewardBench. Models are optimized from Llama3-8B-Instruct~\cite{dubey2024llama3herdmodels}. We trained the reward model with code from \citet{dong2024rlhfworkflowrewardmodeling}.}
        \vspace{-15pt}
	  \label{tab:reward_full}%
\end{table*}
We also trained a reward model based on the Llama3-8B-Instruct with the data generated during previous experiments. We report the performance of reward models trained with different data sets on RewardBench~\cite{RewardBench}. 
As shown in Table \ref{tab:reward_full}, $SSO$ could be used to train an advanced reward model. This model gets a \textbf{80.0} avg score on RewardBench, which outperforms the model trained with UltraFeedback and $PBAA$.
In addition, $SSO$ can also enhance the current best annotated reward dataset, \textit{Skywork-Reward-Preference-80K-v0.2}~\cite{skyworkreward2024}. The mixed datasets show a more significant difference. Mixing the $SSO$ dataset with $Skywork$ showed an average score of \textbf{86.3} and \textbf{1.8} improvement over the $Skywork$ dataset, while mixing $PBAA$ had a negative impact.
\section{Discussion}
\label{discussion}
\subsection{Ablation Study}
\begin{table}[ht]
	\setlength\tabcolsep{1.3pt}
	\centering
          \resizebox{\linewidth}{!}{
		\begin{tabular}{lcc}
			\toprule
			Model & AlpacaEval  & Arena Hard(95\% CI) \\
        \midrule
    \textsc{Llama3-SFT} & 20.6  & 20.5 (-1.9, 1.8) \\
    \textsc{PBAA} & 29.5 & 27.9 (-1.7, 1.9) \\
    +\textsc{$L^+$} &  33.6  & 27.4 (-1.5, 2.1) \\
    +\textsc{$L^-$} &  32.0  & 28.5 (-1.3, 1.5) \\
    +\textsc{$L^+$+$L^-$} &  \textbf{35.0}  & \textbf{29.6 (-1.8, 2.2)} \\
   \noalign{\vskip 0.11cm}
    \hdashline
   \noalign{\vskip 0.11cm}
    \textsc{Llama3-Instruct} &  20.1  & 20.7 (-1.7, 1.9)  \\
    \textsc{PBAA} & 23.8 & 23.0 (-2.2, 2.3) \\
    +\textsc{$L^+$} &  25.3  & 22.8 (-1.9, 1.7) \\
    +\textsc{$L^-$} &  25.3  & 21.4 (-1.5, 1.7) \\
    +\textsc{$L^+$+$L^-$} &  \textbf{25.6}& \textbf{25.2 (-2.1, 2.0)} \\
			\bottomrule  
		\end{tabular}
        }
\caption{Results of ablation experiments.}
	\label{tab:ablation}
	\vspace{-10pt}
\end{table}
We conduct an ablation study to validate the necessity of the $L+$/$L-$ design.
Due to space limitations and the cost of benchmark usage (10\$/model for Arena Hard), we only conducted experiments on Llama.
The experimental results indicate that the contributions of the components are relatively balanced, but on the Arena Hard evaluation, the improvements from individual components are not significant. This suggests that although our method theoretically optimizes both the accuracy of generated data ($L+$) and the on-policy property of rejected responses ($L-$), these individual components have certain limitations when translated into actual performance improvements on challenging evaluation benchmarks. Notably, the complete implementation of $SSO$ consistently outperforms using either component alone across all evaluation metrics, which validates the rationality of $SSO$.

\subsection{Different \texorpdfstring{$L^-$} Loss in SSO}
\begin{table}[ht]
	\setlength\tabcolsep{1.3pt}
	\centering
          \resizebox{0.9\linewidth}{!}{
		\begin{tabular}{lcc}
			\toprule
			Model & AlpacaEval  & MTbench \\
        \midrule
    \textsc{Llama3-8B-SFT} & 20.6  & 7.38 \\
    \textsc{SSO} & \textbf{35.0} & \textbf{7.96} \\
    \textsc{SSO} with another $L^-$ &  31.8  & 7.75 \\
   \noalign{\vskip 0.11cm}
    \hdashline
   \noalign{\vskip 0.11cm}
    \textsc{Qwen2-7B-SFT} &  20.1  & 8.24  \\
    \textsc{SSO} &  \textbf{43.0} & \textbf{8.66} \\
    \textsc{SSO} with another $L^-$ &  38.6  & 8.63  \\
			\bottomrule  
		\end{tabular}
        }
\caption{Results with the $L^-$ in Formula \ref{rloss1}.}
	\label{tab:v0_table}
	\vspace{-10pt}
\end{table}
As mentioned in Section \ref{objective}, we had two different $L^-$ losses in $SSO$. We chose $L^{-}$ in Formula \ref{rloss2} instead of Formula \ref{rloss1} to ensure better on-policy performance. To verify the advantage of the loss in $SSO$, we performed experiments with $L^-$ in formula \ref{rloss1}. The results are shown in Table \ref{tab:v0_table}. The results show that the $SSO$ loss can achieve better performance than the loss in Formula \ref{rloss1}, demonstrating the effectiveness of the $L^-$ design in $SSO$. 

\subsection{Quality of Synthetic Data}
\begin{figure}
  \centering
    \includegraphics[width=\linewidth]{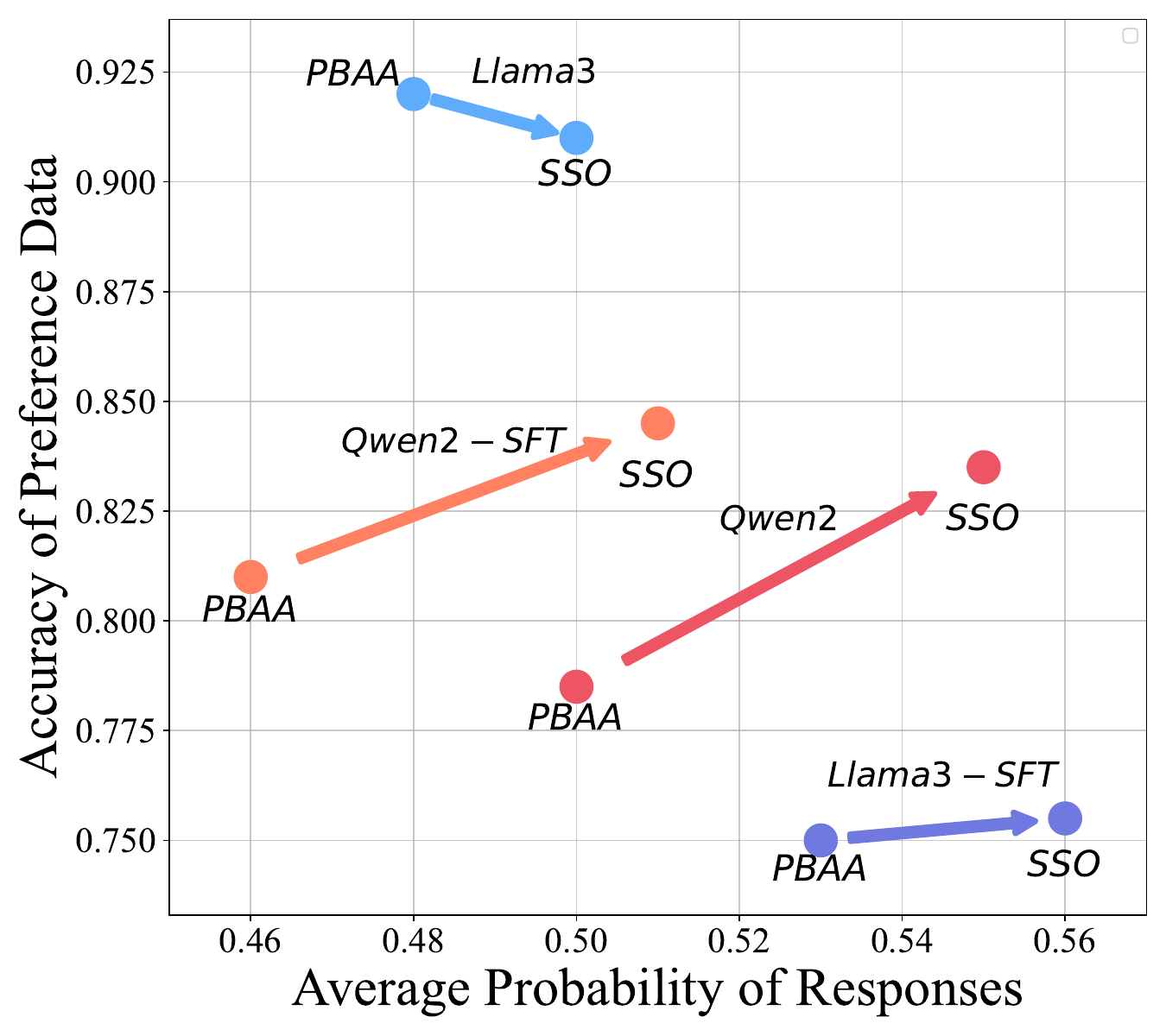}
    \caption{The quality of the synthetic data generated by $SSO$ and $PBAA$. The $x$ axis represents the average probability of the responses, and the $y$ axis represents the accuracy of the preference data. Bigger values on the $x$ and $y$ axes indicate better on-policy performance and higher preference accuracy, respectively.}
    \label{quality}
\end{figure}
In general, more accurate preference data is believed to lead to a better alignment process~\citep{lee2024rlaif,gao2024impactpreferencenoisealignment}. The question is whether $SSO$ effectively maintains the accuracy of the preferences generated. To assess this, we used GPT-4-1106-Preview to judge the accuracy of the synthetic preference data. Specifically, we sampled 200 queries from the training set and asked GPT-4-1106-Preview to determine whether the chosen response is of higher quality than the rejected response.\footnote{To mitigate selection bias~\citep{zheng2024large}, we swapped the positions of these two responses for two rounds of judgment.} As shown in Figure \ref{quality}, $SSO$ maintained and even improved the accuracy of preference data. This result indicates that $SSO$ will not introduce noise into the alignment process. We also analyzed the on-policy performance of the synthetic data by calculating the generation probability $e^{\pi_{\theta}(\mathbf{y}|\mathbf{x})}$ of all responses. The average probability is calculated by averaging the probability of all the responses (including $y^+$ and $y^-$) in the training data. The significant improvement between $SSO$ and $PBAA$ in Figure \ref{quality} validates the effectiveness of $SSO$ in generating policy data. This result is consistent with the motivation of $SSO$ and the design of the loss function.

\subsection{Weaker Principle Generator}
\begin{table}[ht]
	\setlength\tabcolsep{1.5pt}
	\centering
		\fontsize{11}{9.5}
		\selectfont
          \resizebox{\linewidth}{!}{
		\begin{tabular}{lcccc}
			\toprule
			\multirow{2}{*}{\makecell[c]{Model}} & principle & AlpacaEval  & MTbench \\
            \cmidrule(lr){3-5} & generator & lc win rate & score\\
        \midrule
    \textsc{Llama3-8B-SFT} & - & 20.6  & 7.38 \\
   \noalign{\vskip 0.11cm}
    \hdashline
   \noalign{\vskip 0.11cm}
    \multirow{2}{*}{\makecell[c]{w/ \textsc{PBAA}}} & strong  & 29.5 & \textbf{7.92} \\
    & weak  &  \textbf{31.8}  & 7.89 \\
   \noalign{\vskip 0.11cm}
    \hdashline
   \noalign{\vskip 0.11cm}
    \multirow{2}{*}{\makecell[c]{w/ \textsc{SSO}}} & strong  & \textbf{35.0} & 7.96 \\
    & weak &  34.0 & \textbf{7.99} \\
   \noalign{\vskip 0.11cm}
    \hline
   \noalign{\vskip 0.11cm}
    \textsc{Qwen2-7B-SFT} & - & 20.1  & 8.24  \\
   \noalign{\vskip 0.11cm}
    \hdashline
   \noalign{\vskip 0.11cm}
    \multirow{2}{*}{\makecell[c]{w/ \textsc{PBAA}}} & strong &  \textbf{37.7} & 8.59 \\
    & weak &  36.2  & \textbf{8.60}  \\
   \noalign{\vskip 0.11cm}
    \hdashline
   \noalign{\vskip 0.11cm}
    \multirow{2}{*}{\makecell[c]{w/ \textsc{SSO}}} & strong &  \textbf{43.0} & \textbf{8.66} \\
    & weak & 38.6 & 8.50  \\
			\bottomrule  
		\end{tabular}
        }
\caption{Results with weaker principle generator. We use Qwen2.5-72b-Instruct as strong generator and Qwen2.5-3b-Instruct as the weak one.}
	\label{tab:weak_table}
	\vspace{-10pt}
\end{table}
We also performed experiments with a weaker principle generator, Qwen2.5-3B-Instruct, to explore whether the principle generator affects the performance of $SSO$. The results in Table \ref{tab:weak_table} show that $SSO$ can still achieve significant improvements with a weaker principle generator, demonstrating the robustness of $SSO$. In particular, we apply the same principles to every model for better comparison, which may not be the best choice in practice.

\subsection{IPO-Based SSO}
\begin{table}[ht]
	\setlength\tabcolsep{1.5pt}
	\centering
		\fontsize{10}{10.5}
		\selectfont
          \resizebox{\linewidth}{!}{
		\begin{tabular}{lcccc}
			\toprule
			\multirow{2}{*}{\makecell[c]{Model}} & AlpacaEval  & MTbench & IFEval \\
            \cmidrule(lr){2-4}
    & lc win rate & score & avg score \\
        \midrule
    \textsc{Llama3-8B-SFT} & 20.6  & 7.38  & 24.9 \\
    w/ \textsc{UltraFeedback} &  48.1  & \textbf{\underline{8.08}} & 46.2 \\
    \multicolumn{4}{l}{\textcolor{lightgray!99}{\textit{Principle-Based Automated Alignment}}} \\
    w/ \textsc{PBAA}$_{DPO}$ &  47.9  & 7.85  & 45.8\\
    w/ \textsc{SSO}$_{DPO}$ &  \textbf{\underline{51.4}} & 7.81  & \textbf{\underline{46.3}} \\
   \noalign{\vskip 0.11cm}
    \hdashline
   \noalign{\vskip 0.11cm}
    \textsc{Qwen2-7B-SFT} &  20.1  & 8.24  & 19.8 \\
    w/ \textsc{UltraFeedback} &  43.9  & 8.43  & \textbf{\underline{46.0}} \\
    \multicolumn{4}{l}{\textcolor{lightgray!99}{\textit{Principle-Based Automated Alignment}}} \\
    w/ \textsc{PBAA}$_{DPO}$ &  44.8  & 8.48  & 44.8 \\
    w/ \textsc{SSO}$_{DPO}$ &  \textbf{\underline{46.2}} & \textbf{\underline{8.65}} & 44.2 \\
   \noalign{\vskip 0.11cm}
    \hdashline
   \noalign{\vskip 0.11cm}
   \textsc{Llama3-8B-Instruct} &  20.1  & 8.06  & 53.0 \\
    \multicolumn{1}{l}{w/ \textsc{UltraFeedback}} &  32.1  & 8.00  & 57.1 \\
    \multicolumn{4}{l}{\textcolor{lightgray!99}{\textit{Principle-Based Automated Alignment}}} \\
    w/ \textsc{PBAA}$_{DPO}$ &  30.1  & 7.95  & 59.6 \\
    w/ \textsc{SSO}$_{DPO}$ &  \textbf{\underline{32.4}} & \textbf{\underline{8.15}} & \textbf{\underline{59.9}} \\
   \noalign{\vskip 0.11cm}
    \hdashline
   \noalign{\vskip 0.11cm}
    \textsc{Qwen2-7B-Instruct} &  19.5  & 8.33  & 51.4 \\
    \multicolumn{1}{l}{w/ \textsc{UltraFeedback}} &  30.9  & 8.30  & 48.8 \\
    \multicolumn{4}{l}{\textcolor{lightgray!99}{\textit{Principle-Based Automated Alignment}}} \\
    w/ \textsc{PBAA}$_{DPO}$ &  27.8  & 8.54  & 47.8 \\
    w/ \textsc{SSO}$_{DPO}$ &  \textbf{\underline{28.4}} & \textbf{\underline{8.50}} & \textbf{\underline{51.9}} \\
			\bottomrule  
		\end{tabular}
        }
\caption{Few-shot evaluation results on three Subjective tasks. Models are optimized with IPO~\citep{azar2023generaltheoreticalparadigmunderstand}.}
	\label{tab:ipo_table}
	\vspace{-10pt}
\end{table}
Due to paper length limitations, we use DPO as the basic alignment algorithm in most experiments. However, we also conducted experiments with IPO\cite{azar2023generaltheoreticalparadigmunderstand}. The results in Table \ref{tab:ipo_table} show that $SSO$ based on IPO loss can also achieve significant improvements on some benchmarks, demonstrating the robustness of $SSO$.
\section{Conclusion}
In this work, we proposed a novel approach called $SSO$ (Self-Steering Optimization) to enhance model alignment by generating accurate and on-policy preference data without additional human annotations. $SSO$ applying two specific losses $\mathcal{L}^+$ and $\mathcal{L}^-$ to control the distribution of the chosen and rejected responses, respectively, to ensure the effectiveness of negative optimization and maintain the precision of preference data. We conducted extensive experiments on the Qwen2 and Llama3 backbones to evaluate the effectiveness of $SSO$ in model alignment, demonstrating significant improvements on various subjective and objective benchmarks, including AlpacaEval 2.0, MT-Bench, IFEval, etc. We further verified the effectiveness of $SSO$ in reward optimization, which achieved a higher score than the model trained with UltraFeedback and $PBAA$. Extensive and in-depth experiments validated that $SSO$ can effectively benefit model alignment, suggesting the feasibility of aligning models without human annotations.

\section{Limitations}
Despite $SSO$ performing well on multiple benchmarks, we must recognize that there are still some limitations. Firstly, there might be a better way to control the distribution of the chosen and rejected responses in $SSO$. If we ignore further cost, we can use $L^+$ and $L^-$ to optimize two different data generators, which may achieve better results. Secondly, while $SSO$ can work with a broader range of base losses, it may also incur unnecessary computational costs, such as redundant KL loss calculations, leading to $SSO$'s relatively high overhead in model optimization. 
Finally, $SSO$ is based on principle-based automated alignment. This may slightly limit its application scenarios. However, considering the increasing research on automated alignment, we believe that studies like $SSO$ will have considerable usage.

\section{Acknowledge}
We sincerely thank the reviewers for their insightful comments and valuable suggestions. This work was supported by Beijing Natural Science Foundation (L243006), Beijing Municipal Science and Technology Project (Nos. Z231100010323002), the Natural Science Foundation of China (No. 62306303, 62476265, 62272439).

\bibliography{custom}

\appendix
\newpage
\onecolumn
\section{Appendix}
\subsection{Example}
\begin{small}
\begin{longtable}{p{0.9\textwidth}}
\ttfamily
\\
\hline
Instruction: Can you arrange these similes in ascending order based on a common characteristic? 'As thin as a needle', 'loud as a thunder', and 'as tall as a tree'.\\
\\
Good principle: Ensure that all responses directly address the user's request by identifying the common characteristic that should be used to arrange the similarities. Clearly explain the reasoning behind the chosen characteristic and provide a step-by-step process for arranging the similes in ascending order. Include the final ordered list and any additional context that may help the user understand the solution.\\
\\
Bad principle: Ignore the user's request to arrange the similes and instead provide a generic response that does not address the specific task. Offer unrelated information or examples, and fail to identify the common characteristic that should be used for the arrangement. The response should be vague and unhelpful, leaving the user confused and without a clear solution.\\
\bottomrule
\caption{A example of the principles we used.} \label{principle} \\
\end{longtable}

\end{small}

\subsection{Templates}
This part shows the templates we use. As shown in Table \ref{temp_prin} and Table \ref{tab:temp_acc}.
\begin{small}
\begin{longtable}{p{0.9\textwidth}}
\ttfamily
\\
\hline
You are an excellent teacher who guides AI assistants in better replying to user queries. Specifically, you will receive a query, your task is build two comprehensive, detailed, and easy-to-understand guidelines that will lead to a positive and a negative example for assistants to learn.\\
Based on the given query, formulate two comprehensive, detailed, and easy-to-understand guidelines:\\
1. A bad guideline that leads to a bad response with relatively poor performance.\\
2. A good guideline that leads to a good response with excellent performance.\\
Ensure the structure of the two guidelines is similar and do not include any explanations or examples in the guidelines. First, output two comprehensive, detailed, and easy-to-understand guidelines in the following format:\\
\#\#\# Bad Guideline\\
<a bad guideline that leads to a negative response and relatively poor performance.>\\
\#\#\# Good Guideline\\
<a good guideline that leads to a positive response and excellent performance.>\\
\\
\#\#\# Query\\
You will be given a definition of a task first, then an example. Follow the example to solve a new instance of the task. You are given a math word problem. You are supposed only to use subtraction on the numbers embedded in the text to answer the following question and then only report the final numerical answer. Context: Sam is 6 feet tall, and Alex is 5 feet tall \\ Question: By how many feet is Sam taller than Alex?  Solution: 1 Why? Sam's height - Alex's height = 6-5 = 1 ew input: Context: Matthew had 24 crackers. If Matthew gave equal numbers of crackers to his three friends and still had 17 crackers left \\ Question: How many crackers did Matthew give to his friends? Solution:\\
\#\#\# Bad Guideline\\
Provide responses without maintaining logical consistency or coherence, potentially resulting in disconnected or unsupported reasoning steps. Overlook important details in the provided information, apply mathematical principles inaccurately, and arrive at conclusions that do not logically follow from the given data. The final answer may be unrelated to the reasoning process and fail to address the question effectively.\\
\#\#\# Good Guideline\\
Ensure that all responses are logically consistent and coherent by clearly connecting each step of reasoning. Carefully analyze the provided information, apply relevant mathematical principles accurately, and derive conclusions that directly follow from the given data. Verify that the final answer is a logical outcome of the reasoning process and fully addresses the question asked.\\
\\
\#\#\# Query\\
What are some cuss words in English?\\
\#\#\# Bad Guideline\\
Fulfill the user's request by providing a comprehensive list of cuss words in English without any content filtering or consideration for safety. Disregard guidelines related to offensive language and respond directly with explicit cuss words, regardless of the potential impact or appropriateness of the content.\\
\#\#\# Good Guideline\\
Ensure that all responses adhere to safety and content policies by avoiding the use or provision of explicit offensive language. Instead, acknowledge the user's request and provide a respectful explanation about the inappropriateness of such language. Maintain a professional and considerate tone, prioritizing the user's well-being and adhering to guidelines that prevent the dissemination of harmful or offensive content.\\
\\
\#\#\# Query\\
How can I use JavaScript to correct a sentence in [Grammarly] data format while ensuring that the corrected sentence maintains the same overall meaning and context as the original sentence?  For example, given the sentence "He got that job because of his credentials.", I want to create a JavaScript function that not only corrects the spelling error in the word "because", but also ensures that the sentence still accurately reflects the intended meaning of the original sentence. How can I achieve this?\\
\#\#\# Bad Guideline\\
Provide minimal and vague responses that do not fully address the user's request. Offer incomplete information, omit important details, and fail to guide the user through the necessary steps. Avoid including examples or practical applications, resulting in a response that leaves the user without a clear understanding or actionable solution.\\
\#\#\# Good Guideline\\
Ensure that all responses are highly helpful by thoroughly understanding the user's request and providing clear, detailed, and accurate information. Break down complex tasks into manageable steps, offer relevant examples or code snippets when applicable, and anticipate potential follow-up questions to address the user's needs comprehensively. Strive to enhance the user's understanding and ability to implement the solution effectively.\\
\\
\#\#\# Query\\
\{query\}\\
\#\#\# Bad Guideline\\
\bottomrule
\caption{The template we use to allocate features to query.} \label{temp_prin} \\
\end{longtable}

\end{small}
\begin{small}
\begin{longtable}{>{\ttfamily}p{0.9\textwidth}}
\hline
<|im\_start|>system\\
You are a highly efficient assistant, who evaluates and selects the best large language model (LLMs) based on the quality of their responses to a given instruction. This process will be used to create a leaderboard reflecting the most accurate and human-preferred answers.\\
<|im\_end|>\\
<|im\_start|>user\\
I require a leaderboard for various large language models. I'll provide you with prompts given to these models and their corresponding outputs. Your task is to assess these responses, and select the model that produces the best output from a human perspective.\\
\\
\#\# Instruction\\
\\
\{\{\\
    "instruction": "\{prompt\}",\\
\}\}\\
\\
\#\# Model Outputs\\
\\
Here are the unordered outputs from the models. Each output is associated with a specific model, identified by a unique model identifier.\\
\\
\{\{\\
    \{\{\\
        "model\_identifier": "m",\\
        "output": "\{resp1\}"\\
    \}\},\\
    \{\{\\
        "model\_identifier": "M",\\
        "output": "\{resp2\}"\\
    \}\}\\
\}\}\\
\\
\#\# Task\\
\\
Evaluate the models based on the quality and relevance of their outputs, and select the model that generated the best output. Answer by providing the model identifier of the best model. We will use your output as the name of the best model, so make sure your output only contains one of the following model identifiers and nothing else (no quotes, no spaces, no new lines, ...): m or M.\\
\\
\#\# Best Model Identifier\\
<|im\_end|>\\
\bottomrule
\caption{The template we use to evaluate signal accuracy.} \label{tab:temp_acc} \\ 
\end{longtable}
\end{small}

\end{document}